\documentclass{article}

\usepackage[preprint]{aaj2026}

\usepackage[utf8]{inputenc} % allow utf-8 input
\usepackage[T1]{fontenc}    % use 8-bit T1 fonts
\usepackage{hyperref}       % hyperlinks
\usepackage{url}            % simple URL typesetting
\usepackage{booktabs}       % professional-quality tables
\usepackage{amsfonts}       % blackboard math symbols
\usepackage{nicefrac}       % compact symbols for 1/2, etc.
\usepackage{microtype}      % microtypography
\usepackage{xcolor}         % colors
\usepackage{graphicx}
\usepackage{subcaption}

\title{Example-Based Object Detection}

\author{%
    ZhiXin Sun\\
  PowerChina Zhongnan Engineering Corporation Limited \\
  \texttt{sunzxjdi@gmail.com} 
}

\begin{document}

\maketitle

\begin{abstract}
In recent years, object detection has achieved significant progress, especially in the field of open-vocabulary object detection. Unlike traditional methods that rely on predefined categories, open-vocabulary approaches can detect arbitrary objects based on human-provided prompts. With the advancement of prompt-based detection techniques, models such as SAM3 can even outperform some category-specific detectors trained on particular datasets without requiring additional training on those datasets.
However, despite these advancements, false positives and false negatives still occur. In practical engineering applications, persistent misdetections or missed detections of the same object are unacceptable. Yet retraining the model every time such errors occur incurs substantial costs in terms of human effort, computational resources, and time. Therefore, how to leverage existing false positive and false negative samples to prevent such errors from recurring remains a highly challenging and urgent problem.
To address this issue, we propose EBOD (Example-Based Object Detection), 
which integrates a prompt-based detector (SAM3) with robust feature matching 
modules (DINOv3 and LightGlue). The proposed framework effectively suppresses 
the repeated occurrence of false positives and false negatives by leveraging 
previous error examples, without requiring additional model retraining. Code is available at \url{https://github.com/sunzx97/examples_based_object_detection}.
\end{abstract}

\section{Introduction}

Object detection is an important branch of computer vision and plays a crucial role in engineering applications. In recent years, with the development of open-vocabulary object detection\cite{jiang2025detectpointprediction,jiang2024trex2}, it has become possible to rapidly deploy object detection systems in engineering without spending large amounts of time and effort collecting data and training task-specific models. For example, methods such as SAM3\cite{carion2025sam3segmentconcepts} and Grounded SAM\cite{ren2024grounded} only require textual prompts or geometric cues, such as coordinates or bounding boxes, to detect the desired targets.

However, in practical engineering applications, false positives and false negatives are unavoidable due to complex outdoor environments and adverse weather conditions such as fog and heavy rain. Instead of repeatedly collecting misdetection samples and retraining models, 
leveraging false positive and false negative examples to refine detection results 
provides a more efficient alternative, with significantly lower human effort and computational cost.

Recently, several example-based object detection methods have also achieved promising results. SAM3\cite{carion2025sam3segmentconcepts}, T-Rex2\cite{jiang2024trex2} and similar models natively support visual prompts, but these prompts are typically limited to interactive cues such as drawing boxes or adding points within the same image. They do not transfer the representation learned from the bounding box of an object in one image to another image, which would enable few-shot segmentation without relying on text prompts. FSS-SAM3\cite{tsai2026fewshotsemanticsegmentationmeets} addresses this by concatenating the query image and the target image and leveraging SAM3 to enable cross-image visual prompting. However, compared with purely appearance-based visual prompts, SAM3 is more sensitive to geometric cues (e.g., box position and scale), meaning that geometric prompts have a stronger influence on its behavior.
Methods such as INSID3\cite{cuttano2026insid3} and Matcher\cite{liu2023matcher} can detect semantically similar regions in an image using only a small number of examples. Although these semantic-based few-shot detection approaches perform well, they inevitably suffer from semantic recognition errors. In contrast, instance-recognition-based methods\cite{lindenberger2023lightglue,jiang2024Omniglue} are generally more robust, especially since feature point detection and matching models demonstrate strong generalization ability and can achieve excellent performance in unseen scenarios without additional training.
However, example-based matching methods are also affected by differences in scale and field of view between the two images. When the query image is very small and the object occupies only a small number of pixels, while the target image has a much larger field of view and the object occupies only a tiny region, direct feature matching often fails due to insufficient matching keypoints or interference from other regions unrelated to the object in the query image.

Therefore, this paper combines feature matching models such as INSID3\cite{cuttano2026insid3} and LightGlue\cite{lindenberger2023lightglue} with open-vocabulary detectors based on text or box prompts, such as SAM3\cite{carion2025sam3segmentconcepts}, and proposes an example-based object detection algorithm. By providing the model with examples of previous false positives and false negatives, the proposed method avoids repeated detection errors on the same type of example.

\section{Method}
\begin{figure}[htbp]
    \centering
    \includegraphics[width=\linewidth]{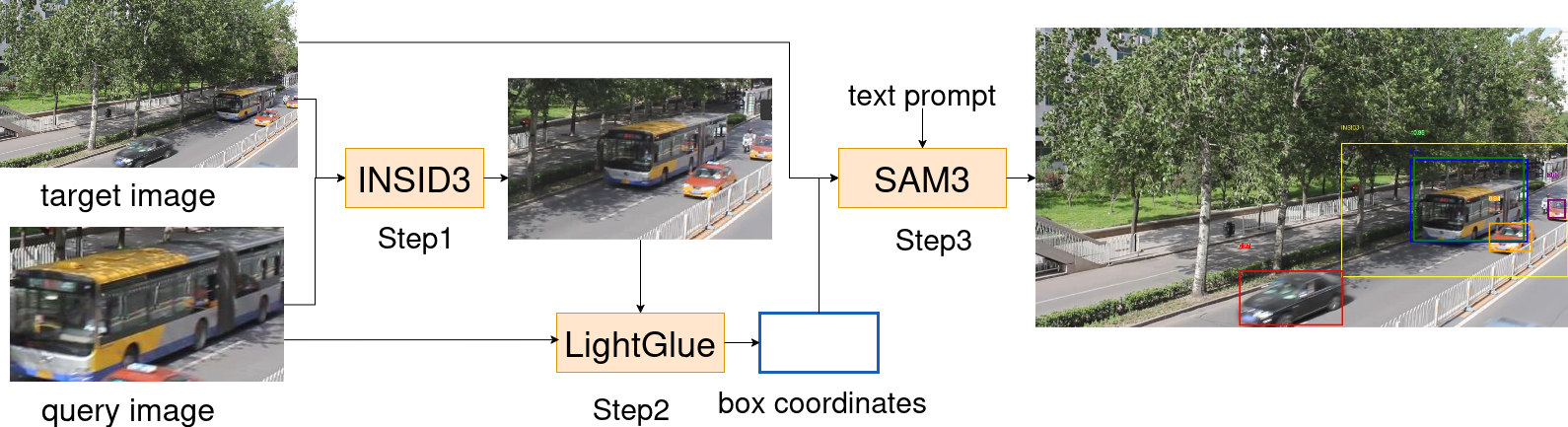}
    \caption{Overview of the proposed EBOD framework. 
    Step1: Use INSID3 to generate candidate regions for the query image, indicated by yellow bounding boxes. 
    Step2: Use LightGlue to match the query image with the candidate boxes generated by INSID3, obtaining precise target box coordinates, shown in blue. 
    Step3: Using the precise box prompt from LightGlue and the user's text prompt, use SAM3 to obtain the detected target boxes, shown in green.}
    \label{figframework}
\end{figure}

The overall architecture of the proposed EBOD framework is illustrated in Fig.~\ref{figframework}. First, INSID3\cite{cuttano2026insid3} is used to search the target image and identify all regions that may contain objects matching the query image. Then, these candidate regions are matched with the query image using LightGlue\cite{lindenberger2023lightglue} for keypoint-based feature matching, producing precise matching bounding boxes. These bounding box coordinates are then used as box prompts, where their labels can be either true or false. Together with the user’s text prompt, they are fed into SAM3\cite{carion2025sam3segmentconcepts} to obtain the final detection results.

% \label{headings}

\subsection{INSID3}

INSID3\cite{cuttano2026insid3} is an in-context segmentation method. Given one annotated visual example, it is a training-free approach that segments concepts at varying granularities using only frozen DINOv3\cite{simeoni2025dinov3} features. Through the robust feature extraction capability of DINOv3\cite{simeoni2025dinov3}, pixel-level feature representations can be obtained.
In this paper, INSID3\cite{cuttano2026insid3} supports both point-based matching and mask-based matching. 
We directly adopt its point-based query mechanism, and the practical pipeline is as follows: 
given a target image and a query image, we first use the backbone model DINOv3\cite{simeoni2025dinov3} in INSID3 to extract the feature of the center point in the query image. This feature is then matched with the features of all pixels in the target image. A threshold $\sigma$ is applied to filter out low-confidence responses. For the high-confidence pixels, DBSCAN\cite{10.5555/3001460.3001507} clustering is performed with the minimum number of samples set to $3$. The bounding box of each cluster is extracted and further expanded according to the size of the query image. Specifically, let $w$ and $h$ 
denote the width and height of the query image, and let the boundary of a cluster be 
$(x_{\min}, y_{\min}, x_{\max}, y_{\max})$. After expansion, the candidate bounding box becomes 
$(x_{\min}-w/2,\; y_{\min}-h/2,\; x_{\max}+w/2,\; y_{\max}+h/2)$.

\subsection{LightGlue}
LightGlue\cite{lindenberger2023lightglue} is an image matching model that integrates keypoint detection and feature matching into a unified framework. Given two input images, LightGlue\cite{lindenberger2023lightglue} first employs feature extractors such as SuperPoint\cite{detone2018superpointselfsupervisedpointdetection} to detect keypoints and extract local descriptors. It then performs feature correspondence matching between the two images through a dedicated matching network. According to the matching difficulty between the image pair, LightGlue\cite{lindenberger2023lightglue} can adaptively terminate at different network layers and output the matching results efficiently.

In this paper, LightGlue\cite{lindenberger2023lightglue} is applied to perform feature matching between the query image and the candidate regions generated by INSID3\cite{cuttano2026insid3}. Based on the matched feature points, a homography matrix $H$ is estimated. Then, the coordinates of the four vertices of the query image are projected onto the candidate region according to the homography transformation. Let the four corners of the query image be defined as

\[
(0,0),\ (w_q,0),\ (w_q,h_q),\ (0,h_q),
\]

where $w_q$ and $h_q$ denote the width and height of the query image, respectively. The transformed corner coordinates in the candidate region are obtained by

\[
\mathbf{p}_i' = H\mathbf{p}_i,\quad i=1,2,3,4,
\]

where $\mathbf{p}_i$ are the homogeneous coordinates of the query-image corners, and $\mathbf{p}_i'$ are the corresponding projected coordinates.

Afterward, by adding the top-left coordinate offset $(x_c,y_c)$ of the candidate bounding box in the original target image, the final coordinates of the matched object in the target image can be obtained as

\[
\mathbf{\hat p}_i = \mathbf{p}_i' + (x_c,y_c),\quad i=1,2,3,4.
\]

To filter out incorrect matches, a candidate region is considered a valid match only when the number of matched keypoints exceeds a predefined threshold and the inlier ratio is above a certain proportion. Here, inliers refer to the matched points that satisfy the same homography matrix $H$.

\subsection{SAM3}

SAM3\cite{carion2025sam3segmentconcepts} is a unified model that integrates detection, segmentation, and tracking. Owing to its excellent generalization capability, it is adopted in this paper as the object detector. Given a text prompt, SAM3\cite{carion2025sam3segmentconcepts} can detect the corresponding target objects in the input image. Compared with fixed-category detection models such as YOLO\cite{redmon2016you}, SAM3\cite{carion2025sam3segmentconcepts} additionally supports visual prompts, including bounding boxes and point annotations.

To avoid repeated false negatives and false positives of the same instances, we first collect representative missed and incorrectly detected targets as query examples. Then, INSID3\cite{cuttano2026insid3} is employed to roughly estimate their locations in the target image and generate candidate regions. After that, LightGlue\cite{lindenberger2023lightglue} is applied to perform feature matching between each candidate region and the corresponding query image, producing accurate coordinate box prompts. Finally, these refined box prompts are fed into SAM3\cite{carion2025sam3segmentconcepts} together with the text prompt for detection.
This pipeline effectively suppresses the repeated occurrence of the same false-positive and false-negative objects.

Compared with purely semantic-based or instance-matching approaches, 
EBOD combines coarse semantic retrieval with precise geometric verification, 
resulting in improved robustness under viewpoint and scale variations.

\section{Experiments}
We evaluate the proposed method in real-world scenarios, 
where it demonstrates strong effectiveness in reducing repeated false positives 
and false negatives for the same object instances. Due to data privacy constraints, we are unable to release additional details, Only one example is provided here to illustrate the process.

\begin{figure}[htbp]
    \centering
    \begin{subfigure}[b]{0.45\textwidth}
        \centering
        \includegraphics[width=\textwidth]{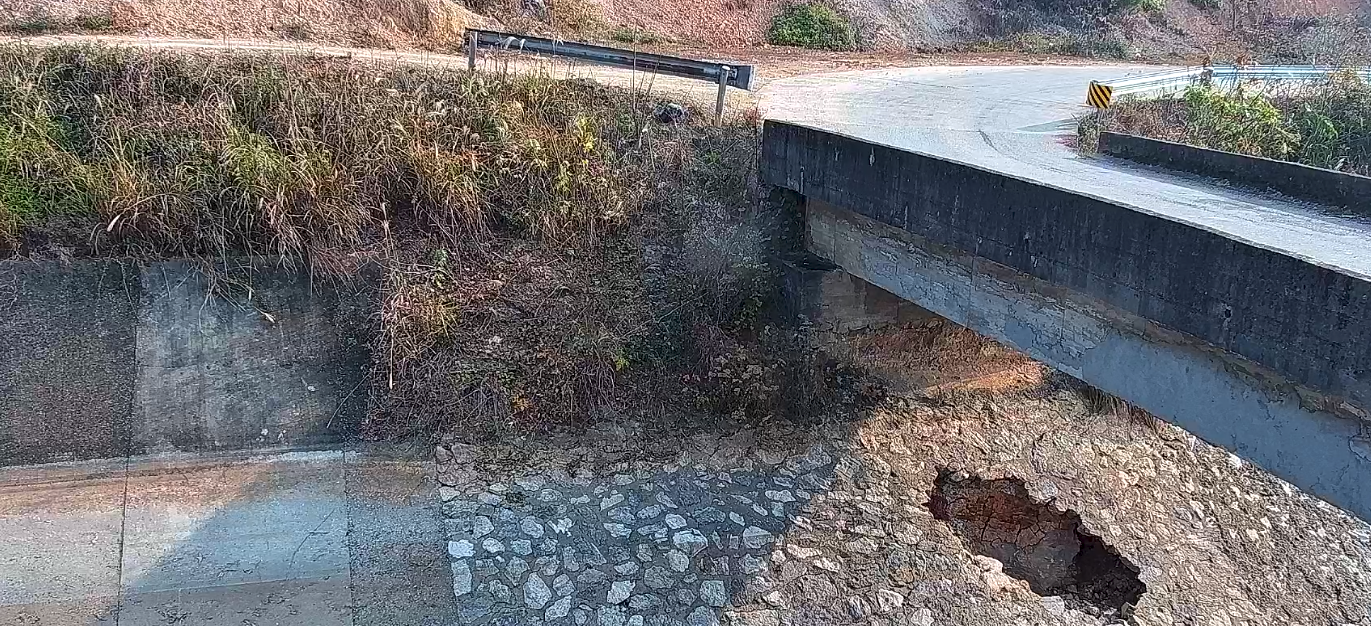}
        \caption{Viewpoint 1}
        \label{fig:img1}
    \end{subfigure}
    \hfill
    \begin{subfigure}[b]{0.45\textwidth}
        \centering
        \includegraphics[width=\textwidth]{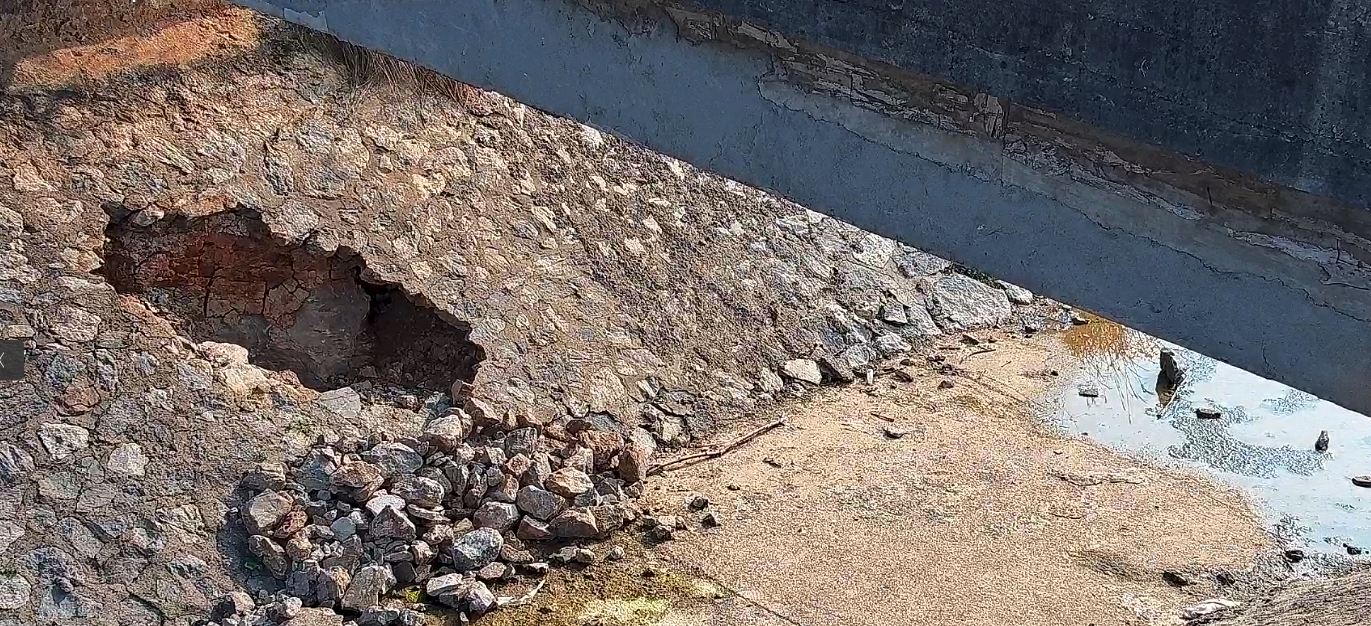}
        \caption{Viewpoint 2}
        \label{fig:img2}
    \end{subfigure}
    \caption{Images of the same object from two different viewpoints}
    \label{figtwoview}
\end{figure}

\begin{figure}[htbp]
    \centering
    \includegraphics[width=\linewidth]{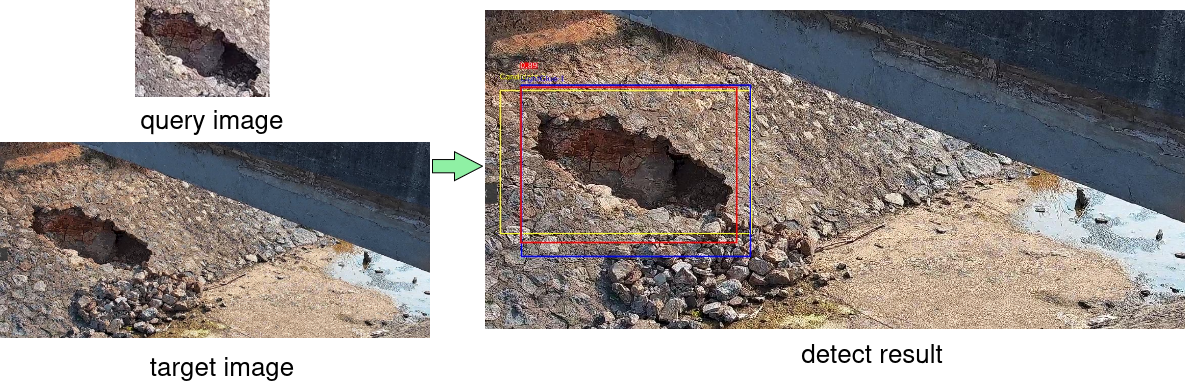}
    \caption{Given a missed detection case, we visualize the detection results produced by the proposed EBOD framework, where the yellow boxes correspond to candidate regions generated by INSID3, the blue boxes denote refined regions obtained via LightGlue-based matching, and the green boxes represent the final detection results of SAM3 conditioned on box prompts.}
    \label{figresult}
\end{figure}

As shown in the Fig.~\ref{figtwoview}, the same collapse appears in images captured from two different viewpoints. We first take the collapsed region in the image from Viewpoint 1 as the query image, and then use the proposed method to detect the corresponding target in the image from Viewpoint 2. Fig.~\ref{figresult} shows the detection results, where the yellow boxes indicate candidate regions generated by INSID3\cite{cuttano2026insid3}, the blue boxes indicate refined regions selected by LightGlue\cite{lindenberger2023lightglue} matching, and the green boxes indicate the final detection results of SAM3\cite{carion2025sam3segmentconcepts} under the box prompt.

The experimental results demonstrate that, although the images are captured from different viewpoints, the method successfully prevents the same missed detection from occurring again.

However, we also observed during experiments that INSID3\cite{cuttano2026insid3} seems to become ineffective for objects occupying only a very small number of pixels. How to reliably generate coarse candidate regions regardless of object size remains a challenging problem.

\section{Conclusion}
In this paper, we propose EBOD, an example-based object detection framework 
that leverages historical false positive and false negative samples to 
improve detection robustness without retraining. By integrating INSID3 for 
coarse candidate region generation, LightGlue for precise instance-level 
matching, and SAM3 for prompt-based detection, the proposed method effectively 
reduces repeated detection errors in real-world engineering scenarios.
Experimental results demonstrate that EBOD can successfully transfer 
object-level information across different viewpoints and prevent recurring 
missed detections. This makes it particularly suitable for practical deployments 
where rapid iteration and low annotation cost are critical.
Despite its effectiveness, the current framework still relies on INSID3 for 
initial candidate generation, which may become unreliable when the target 
object occupies only a very small region in the image. In future work, we plan 
to explore more robust candidate proposal strategies that are less sensitive 
to object scale, as well as more efficient mechanisms for filtering noisy 
matches.

\bibliography{biblio.bib}

\end{document}